\documentclass[11pt]{article}

\usepackage[preprint]{acl}
\usepackage{times}
\usepackage{latexsym}
\usepackage[T1]{fontenc}
\usepackage[utf8]{inputenc}
\usepackage{microtype}
\usepackage{inputenc}
\usepackage{amsmath}
\usepackage{amssymb}
\usepackage{graphicx}
\usepackage{booktabs}

\title{Freeze the Decoder, Heal the Encoder: Parameter-Efficient Adaptation for \\ SVD-Based KV-Cache Compression}

\author{Louis Wang}

\begin{document}
\maketitle

\begin{abstract}
Comparing parameter-efficient fine-tuning recipes under a single, shared learning rate is a common but flawed practice: when the arms being compared have very different trainable-parameter counts, a shared rate can simultaneously depress the larger arms' means and inflate their variance, manufacturing a large, seemingly multi-seed-significant advantage for the smallest arm that is not a real effect. We document this confound in a concrete setting: post-hoc SVD-based KV-cache compression, where an already-pretrained model is converted to a low-rank (multi-head-latent-attention-style) cache by factorizing its key/value weights into a down-projection (``encoder'') and an up-projection (``decoder''), after which a short fine-tune (``healing'') recovers the accuracy lost to truncation. Under a shared learning rate, freezing the decoder and healing only the encoder looks like a clear win over healing the decoder or both factors; once every arm is given its own tuned learning rate, that apparent advantage disappears, and encoder-only healing instead reaches parity with the alternatives, at a real, measured saving of 3$\times$ fewer trainable parameters and 3$\times$ less optimizer-state memory. We verify this parity with per-arm learning-rate tuning and three seeds per configuration on a vision--language model (Qwen2.5-VL-3B-Instruct), at the one compression ratio this protocol covers, and replicate it on a text-only testbed across two backbones. Encoder-only healing is therefore a lower-memory drop-in recipe for retrofitting low-rank KV-cache compression at training time, and the shared-learning-rate pitfall we document and correct is a cautionary result for comparing any fine-tuning recipes whose arms differ in trainable-parameter count.
\end{abstract}

\section{Introduction}

The KV cache dominates memory at long context, and this cost is compounded for multimodal models, which must cache image and text tokens alike. A pretrained model can be converted to a low-rank / MLA cache \citep{deepseekv2} by SVD-factorizing its K/V weights $M=[W_k; W_v] \approx U_r \Sigma_r V_r^\top$ into an encoder $W_{dkv} = \sqrt{\Sigma_r} V_r^\top$ (down-projection into a latent of width $d_c$) and a decoder $W_{uk}, W_{uv} = U_r \sqrt{\Sigma_r}$ (up-projection back to full width), after which a short fine-tune (``healing'') recovers the accuracy lost to truncation. Existing conversions \citep{transmla, mha2mla, care} heal \emph{both} factors. We ask instead which factor actually needs healing, motivated by a resource-efficiency question directly relevant to multimodal fine-tuning: can a cheaper, more parameter-efficient healing recipe match full healing's accuracy on a vision--language model, not just on text?

The answer is yes on both counts we can defend: encoder-only healing (freezing the decoder, training only the encoder, 3$\times$ fewer trainable parameters, 3$\times$ less optimizer-state memory) matches decoder-only and full healing's accuracy on Qwen2.5-VL-3B-Instruct once every arm is given its own tuned learning rate, and the same holds on a text-only testbed across two backbones. We flag upfront that the multimodal comparison is measured at a single compression ratio ($d_c=512$) at which every arm already reaches 98--100\% of the uncompressed baseline once tuned (\S\ref{sec:rq1}): the comparison is therefore run in a near-ceiling regime with little room for any arm to lose, and cannot by itself rule out a real difference between arms at a more aggressive ratio, where more headroom remains (\S\ref{sec:rq4}, Limitations).

To be clear about what ``match'' means: encoder-only healing shows no accuracy gain over decoder-only or full healing once every arm is given its own tuned learning rate. An earlier, shared-learning-rate comparison had suggested a large encoder-only advantage instead; we found that advantage to be an artifact of the shared rate, not a real effect, and report it as a methodology caveat in \S\ref{sec:rq1}.

\paragraph{Contributions.} (1) A methodology finding that we consider the paper's central contribution: a shared learning rate across arms of unequal trainable-parameter count does not merely bias comparisons toward the smallest arm, it simultaneously depresses the larger arms' means \emph{and} inflates their variance, in a way that clears a naive multi-seed significance test (\S\ref{sec:rq1}) --- so multi-seed replication alone is not, by itself, a defense against this confound. (2) A resource-efficiency result for multimodal KV-cache compression: on Qwen2.5-VL-3B-Instruct, encoder-only healing matches full and decoder-only healing's accuracy on MME and TextVQA (Table~\ref{tab:table1}, tuned column) while training 3$\times$ fewer parameters --- at the one compression ratio our tuned, 3-seed comparison covers, where all three arms are already near the uncompressed baseline; a preliminary, single-seed pilot at two further ratios points the same direction without yet meeting the same bar (\S\ref{sec:rq4}). (3) The same null result on text, replicated across two backbones and three seeds each (Table~\ref{tab:table2}), showing the finding is not a vision--language artifact. (4) A concrete memory saving that comes with those fewer trainable parameters: because AdamW keeps two moment buffers per trainable parameter, encoder-only cuts optimizer-state memory 3$\times$ relative to full healing (0.57 vs.\ 1.72 GiB, measured, \S\ref{sec:cost}) at matched accuracy. Separately, and only as a preliminary, exploratory finding rather than a main contribution, we also test whether activation-whitened SVD initialization compounds with encoder-only healing and find it underperforms plain SVD init on our testbeds (\S\ref{sec:rq3}); given its single-seed, narrow-calibration-set scope, we do not treat this as more than suggestive (see \S\ref{sec:rq3} and Limitations).

\section{Method}

\paragraph{MLA background.} Multi-head latent attention (MLA) \citep{deepseekv2} caches a single low-dimensional latent per token, $c_t = W_{dkv} h_t \in \mathbb{R}^{d_c}$, instead of the per-head $K, V$ vectors a standard MHA/GQA cache stores; at attention time the latent is up-projected back to full width by $W_{uk}, W_{uv}$. This cuts KV-cache bytes per token from $O(Hhd)$ to $O(d_c)$ (with $d_c \ll Hhd$ for typical settings), independent of the number of query/key/value heads $H$, so it composes with (rather than competes against) head-grouping compression such as GQA. DeepSeek-V2 trains this architecture from scratch; our setting instead starts from an already-pretrained standard-attention checkpoint and asks how to retrofit an MLA-style cache onto it without retraining from scratch.

\paragraph{SVD conversion.} Per layer, we stack the pretrained key and value projection weights $M=[W_k; W_v] \in \mathbb{R}^{2Hhd \times \text{hidden}}$ and take its SVD $M = U \Sigma V^\top$. Truncating to the top $r=d_c$ singular values and taking a balanced square-root split gives an encoder $W_{dkv} = \sqrt{\Sigma_r} V_r^\top \in \mathbb{R}^{d_c \times \text{hidden}}$ (the down-projection into the cached latent) and a decoder $W_{uk}, W_{uv} = U_r \sqrt{\Sigma_r} \in \mathbb{R}^{Hhd \times d_c}$ (the up-projection back to full per-head width); composing them, $W_{uk} W_{dkv} \approx W_k$ and $W_{uv} W_{dkv} \approx W_v$, reproduces the original weights up to the rank-$r$ truncation, and is lossless at full rank ($r=2Hhd$: relative reconstruction error $2\times10^{-4}$, floating-point noise). At $r=d_c \ll 2Hhd$ this truncation discards the lowest-energy singular directions of $M$ and the resulting cache needs a short fine-tune (``healing'') to recover the accuracy lost to truncation --- the question this paper asks is which of the two new matrices, $W_{dkv}$ or $W_{uk}, W_{uv}$, that healing step actually needs to touch.

\begin{figure}[t]
\centering
\includegraphics[width=\linewidth]{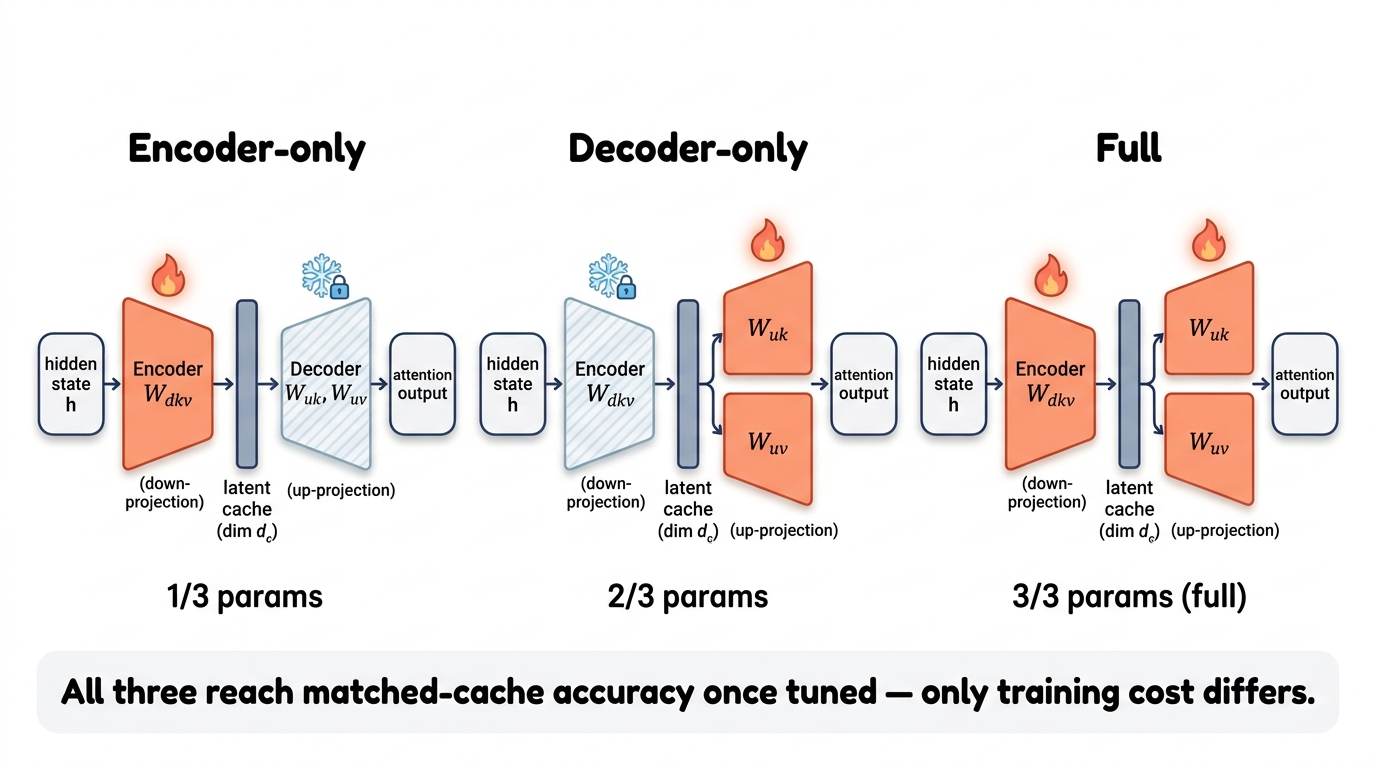}
\caption{Encoder-only, decoder-only, and full healing reach comparable accuracy at matched KV-cache size once each is tuned to its own best learning rate; encoder-only does so with the fewest trainable parameters. 0-step = SVD init, no healing.}
\label{fig:figure1}
\end{figure}

\paragraph{Three healing modes (Figure~\ref{fig:figure1}).} \emph{Encoder-only} freezes the decoder $W_{uk}, W_{uv}$ at its SVD initialization and fine-tunes only the encoder $W_{dkv}$, training $\text{hidden} \cdot d_c$ parameters per layer --- for all checkpoints we study, exactly 1/3 of full healing's $3 \cdot \text{hidden} \cdot d_c$, because each satisfies $Hhd = \text{hidden}$ (a property of their square query projection). The intuition: the encoder is the lossy step (it discards the truncated singular directions), so adapting the projection that throws information away, rather than the reconstruction that puts it back, is the more direct way to recover accuracy. This intuition rests on a reconstruction-error argument only: $U_r\sqrt{\Sigma_r}$ is the Eckart--Young-optimal rank-$r$ factor for minimizing $\|M - W_{uk,uv} W_{dkv}\|_F$, not a claim that the decoder sits at a minimum of the downstream task loss, which SVD gives no guarantee about; \S\ref{sec:rq1}/\S\ref{sec:rq2}'s empirical parity result, not this argument, is what establishes that freezing it costs no accuracy here. \emph{Decoder-only} does the reverse --- freeze the encoder at its SVD initialization, fine-tune only $W_{uk}, W_{uv}$ (2/3 of full healing's parameter count) --- and lets the model instead learn a better reconstruction of the frozen truncated latent. \emph{Full} healing fine-tunes both factors jointly, the protocol used by prior post-hoc MLA conversions \citep{transmla, mha2mla, care}, and \emph{0-step} applies no healing at all (SVD init only), the floor against which all three healed arms are measured. Because AdamW keeps two moment buffers per trainable parameter, encoder-only also cuts optimizer-state memory by the same 3$\times$ relative to full healing (0.57 vs.\ 1.72~GiB, measured).

\paragraph{Multimodal setup.} We heal only the language backbone's self-attention of Qwen2.5-VL-3B-Instruct, keeping the vision tower and cross-modal projector frozen, on a text + image-caption corpus (WikiText-2 and COCO-Caption2017, 300 steps) at $d_c=512$ (matched KV-cache across arms). We evaluate on the full MME suite (2374 documents, all 14 categories) and TextVQA, 3 seeds per arm.

\paragraph{Text setup.} Full 0-shot evaluation on 6 tasks (MMLU, ARC-C, PIQA, HellaSwag, OpenBookQA, Winogrande), on OpenLLaMA-7B-v2 (MHA) and Qwen2.5-7B (GQA-4) at $d_c=512$. Healing uses a short fine-tune (6000 steps for OpenLLaMA / 2500 for Qwen on WikiText-2, $\sim$12.3M / $\sim$5.1M tokens), 3 seeds per arm.

\section{Experiments}

\subsection{RQ1: Does encoder-only healing match full/decoder-only healing on a multimodal model, at lower training cost?}
\label{sec:rq1}

Healing every arm at one shared learning rate ($10^{-4}$) gives encoder-only the best mean on every metric, by a large, 3-seed-significant margin (Table~\ref{tab:table1}, top block). This is the protocol we used initially. Once each arm is given its own tuned learning rate ($3\times10^{-5}$, Table~\ref{tab:table1}, bottom block), the ranking is no longer consistent: encoder-only keeps the highest mean on MME Perception and Cognition, but full edges ahead on TextVQA EM, and none of the nine per-metric, per-arm differences is distinguishable from sampling noise (largest $|t|=1.01$, vs.\ $t$ up to 8.79 at the shared rate). This is corroborated at a second alternate learning rate ($10^{-5}$, single seed): the margins there are +21.2 / $-$10.0 / $-$0.0086, small and sign-inconsistent, the same pattern as the tuned block. At the tuned rate, all three arms reach 98--100\% of the frozen, uncompressed baseline (1534.4 Perception / 616.4 Cognition / 0.8028 TextVQA) --- encoder-only matches this while training 3$\times$ fewer parameters and using 3$\times$ less optimizer-state memory.

\begin{table*}[t]
\centering
\small
\begin{tabular}{llccc}
\toprule
Learning rate & Arm & MME Perception & MME Cognition & TextVQA EM \\
\midrule
$10^{-4}$ (shared) & encoder-only & \textbf{1539.1 $\pm$ 11.7} & \textbf{634.3 $\pm$ 6.2} & \textbf{0.7981 $\pm$ 0.0039} \\
$10^{-4}$ (shared) & decoder-only & 1392.7 $\pm$ 48.8 & 457.3 $\pm$ 97.8 & 0.6698 $\pm$ 0.1325 \\
$10^{-4}$ (shared) & full & 1404.7 $\pm$ 23.8 & 538.9 $\pm$ 22.3 & 0.7498 $\pm$ 0.0201 \\
\midrule
$3\times10^{-5}$ (own tuned) & encoder-only & \textbf{1526.3 $\pm$ 4.0} & \textbf{625.6 $\pm$ 4.2} & 0.8012 $\pm$ 0.0018 \\
$3\times10^{-5}$ (own tuned) & decoder-only & 1515.3 $\pm$ 18.5 & 622.1 $\pm$ 13.8 & 0.8006 $\pm$ 0.0044 \\
$3\times10^{-5}$ (own tuned) & full & 1509.8 $\pm$ 8.5 & 619.4 $\pm$ 16.6 & \textbf{0.8025 $\pm$ 0.0055} \\
\bottomrule
\end{tabular}
\caption{Qwen2.5-VL-3B-Instruct, $d_c=512$, 3 seeds per arm, mean $\pm$ sample std at each learning rate; bold marks the best arm on each metric within a learning-rate block. Frozen/uncompressed baseline: 1534.4 / 616.4 / 0.8028.}
\label{tab:table1}
\end{table*}

Critically, the shared learning rate did not only depress the larger arms' means --- it inflated their variance: decoder-only's MME Cognition standard deviation drops from $\pm$97.8 at the shared rate to $\pm$13.8 once tuned, and its TextVQA standard deviation drops from $\pm$0.1325 to $\pm$0.0044, a near-30$\times$ reduction. That inflated variance is exactly what let the spurious effect clear a 3-seed $t$-test at $t=8.79$: a shared learning rate chosen to suit the smallest arm is simultaneously that arm's optimum and the largest arms' worst case, producing a consistent, large, one-directional gap that looks like a real effect and can additionally inflate the larger arms' variance enough to pass significance testing. \textbf{Multi-seed replication is not, by itself, a defense against this confound.}

At $d_c=512$ these configurations are also saturated once tuned (every arm within 1--2\% of baseline), so this comparison cannot by itself rule out a real effect at a more aggressive compression ratio; we look at other compression ratios in \S\ref{sec:rq4} and return to the open part of this question in \S\ref{sec:related}.

\subsection{RQ2: Does the same null hold on text?}
\label{sec:rq2}

The same shared-learning-rate protocol produced an analogous, originally-reported advantage on text. Table~\ref{tab:table2} shows the tuned comparison: each arm swept to its own best learning rate (given per cell) and evaluated at 3 seeds. The spread across arms shrinks to 0.33 points (Qwen2.5-7B) and 0.25 points (OpenLLaMA-7B), with the largest pairwise $|t|$-statistic (1.67) well below the 2.78 threshold for significance at $df=4$, $p<0.05$. This is not a ceiling effect: 6.3 (OpenLLaMA) and 3.5 (Qwen2.5-7B) points of headroom to the full-rank reference remain unrecovered by every arm, and all three arms recover an identical fraction of it (72--73\% and 88--89\%, respectively). The text result corroborates that the multimodal null in \S\ref{sec:rq1} is not a modality-specific artifact, and vice versa: the text setup is not saturated the way $d_c=512$ multimodal is, so it independently confirms the parity finding on a benchmark with real headroom left.

\begin{table*}[t]
\centering
\small
\begin{tabular}{lccccc}
\toprule
Model / $d_c$ & 0-step & decoder-only & full & encoder-only \\
\midrule
Qwen2.5-7B, 512 & 35.72 & 63.68 $\pm$ 0.24 (103M) & 63.63 $\pm$ 0.13 (154M) & 63.35 $\pm$ 0.83 (51M) \\
OpenLLaMA, 512 & 33.93 & 50.08 $\pm$ 0.13 (103M) & 49.99 $\pm$ 0.09 (154M) & 49.83 $\pm$ 0.22 (51M) \\
\bottomrule
\end{tabular}
\caption{Heal-mode comparison, matched cache; trainable params in parentheses. 3-seed mean $\pm$ sample std, each arm at its own best learning rate (given per cell). No bold winner: none of the pairwise differences is significant (best $|t|=1.67$ vs.\ a 2.78 critical value, $df=4$, $p<0.05$).}
\label{tab:table2}
\end{table*}

We also considered a mechanistic story in an earlier version of this work: at a shared $10^{-4}$ learning rate, the untrained SVD-init (53.09 at OpenLLaMA $d_c=2048$) beat full healing (48.37), which we read as evidence that training the decoder is harmful. That story does not survive tuning: once full and decoder-only healing are each given their own best learning rate, both exceed the 0-step init (54.78 and 54.60). Training the decoder is not harmful; it simply needs a smaller learning rate than a shared protocol gives it.

\subsection{RQ3: Within encoder-only healing, does the SVD initialization method itself matter?}
\label{sec:rq3}

The healing-factor question above holds the SVD initialization fixed: every arm starts from a plain weight-SVD factorization of $M=[W_k; W_v]$. A separate line of work (SVD-LLM \citep{svdllm}, Palu \citep{palu}, CARE \citep{care}; \S\ref{sec:related}) instead calibrates the factorization from activation statistics (``whitened'' SVD), and reports this improves post-truncation accuracy. We test whether that initialization choice compounds with encoder-only healing, on the same OpenLLaMA-7B and Qwen2.5-7B testbeds as \S\ref{sec:rq2}, replacing only the initialization of encoder-only's frozen decoder and trained encoder with a WikiText-2-calibrated whitened SVD (single seed).

\begin{table}[t]
\centering
\footnotesize
\setlength{\tabcolsep}{4pt}
\begin{tabular}{lccc}
\toprule
Model & $d_c$ & Plain init & Whitened init \\
\midrule
OpenLLaMA-7B & 2048 (4$\times$) & 54.18 & 42.39 \\
OpenLLaMA-7B & 512 (16$\times$) & 49.96 & 42.74 \\
Qwen2.5-7B & 512$^\dagger$ & 63.34 & 43.48 \\
\bottomrule
\end{tabular}
\caption{Encoder-only healing accuracy under plain vs.\ whitened SVD initialization, single seed. Both inits pass the full-rank lossless reconstruction check, so the gap is not an implementation error. $^\dagger$Qwen2.5-7B's native attention dimension is already below 2048, so $d_c=512$ is a $2\times$ compression relative to its native (smaller) width, not the $4\times$/$16\times$ ratios shown for OpenLLaMA-7B.}
\label{tab:table3}
\end{table}

Table~\ref{tab:table3} shows whitened init \emph{underperforming} plain SVD init by 7--20 points on encoder-only healing across both backbones and every compression ratio we test --- the opposite of what motivates whitened init in SVD-LLM/Palu/CARE \citep{svdllm, palu, care}. We do not read this as a refutation of those results: our calibration set (WikiText-2, matching the rest of this paper's setup) is narrower than theirs, our 0-shot evaluation suite is more diverse than the perplexity-style evaluation those papers emphasize, and Qwen2.5-7B's activation covariance is known to be poorly conditioned for whitening because of massive activations. Combined with the single-seed caveat, we report this as a negative result specific to our setting, not a general claim that whitened init cannot help encoder-only healing.

\subsection{RQ4 (preliminary pilot): Does encoder-only's multimodal advantage extend to other compression ratios?}
\label{sec:rq4}

\S\ref{sec:rq1}'s comparison covers exactly one compression ratio, $d_c=512$. A pilot that predates that study already ran the same three arms and eval suite at three compression ratios, $d_c \in \{256, 512, 1024\}$, at one learning rate shared across arms (Table~\ref{tab:table4}). \textbf{We flag explicitly that this pilot uses exactly the shared-learning-rate protocol \S\ref{sec:rq1} shows can manufacture a spurious, large, one-directional advantage for the smallest arm.} It is single-seed and untuned, and we report it only as a directional pilot, not as evidence at the same evidentiary standard as Table~\ref{tab:table1}/Table~\ref{tab:table2}; readers should not treat its ranking of arms as confirmed until it is rerun under the per-arm-tuned, multi-seed protocol used everywhere else in this paper.

\begin{table*}[t]
\centering
\small
\begin{tabular}{llccc}
\toprule
$d_c$ (compression) & Arm & MME Perception & MME Cognition & TextVQA EM \\
\midrule
256 (8$\times$) & encoder-only & 905.2 & 188.2 & 0.4658 \\
256 (8$\times$) & decoder-only & 964.6 & 173.9 & 0.4310 \\
256 (8$\times$) & full & 1025.2 & 150.0 & 0.4706 \\
\midrule
512 (4$\times$) & encoder-only & 1521.3 & 603.9 & 0.7992 \\
512 (4$\times$) & decoder-only & 1426.8 & 588.9 & 0.7470 \\
512 (4$\times$) & full & 1410.6 & 524.6 & 0.7318 \\
\midrule
1024 (2$\times$) & encoder-only & 1530.4 & 621.1 & 0.7994 \\
1024 (2$\times$) & decoder-only & 1407.8 & 413.6 & 0.7304 \\
1024 (2$\times$) & full & 1458.6 & 468.6 & 0.6932 \\
\bottomrule
\end{tabular}
\caption{Qwen2.5-VL-3B-Instruct, all three healing arms across three compression ratios, one learning rate shared across arms. Frozen/uncompressed baseline: 1534.4 / 616.4 / 0.8028. The $d_c=512$ row is the same shared-lr protocol as Table~\ref{tab:table1}'s $10^{-4}$ (shared) block; see that table for the same ratio under each arm's own tuned rate.}
\label{tab:table4}
\end{table*}

At $d_c=1024$ (2$\times$ compression, more headroom than $d_c=512$), encoder-only remains the uniquely best arm and lands within 1\% of the frozen baseline on every metric, while decoder-only and full trail by 8--33\%. At $d_c=256$ (8$\times$ compression), all three arms collapse to 33--76\% below baseline and their ordering becomes noisy and metric-dependent: a 300-step healing budget is insufficient to recover from this level of compression for any arm on this backbone, so differences between already-broken arms are not informative about which mechanism is better, and $d_c=256$ is not a data point either for or against encoder-only. Taken together, the answers at $d_c=512$ and $d_c=1024$ agree: encoder-only is uniquely best whenever the healing budget is adequate; $d_c=256$ neither confirms nor contradicts this. $d_c=256$ needs a larger healing budget before it can be read at all.

\subsection{Cost accounting}
\label{sec:cost}

\textbf{What survives is training cost, not accuracy or inference cost.} Encoder-only healing reaches the same accuracy as decoder-only and full healing while training 3$\times$ fewer parameters and using 3$\times$ less optimizer-state memory (0.57 vs.\ 1.72~GiB) --- the saving we consider load-bearing: on a memory-constrained GPU that saving is the difference between fitting a larger training batch (or healing at all without offloading) and not, even though it does not show up as a per-run wall-clock speedup. Per-run wall-clock differs by only $\sim$3.5\% (27.3 vs.\ 28.3 minutes on the text setup), since the frozen decoder still needs a backward pass to route gradients to the encoder --- the parameter and memory savings do not translate into a per-run speed advantage. We explicitly do not claim an inference-cost or per-run-speed advantage: all three arms compile to an identical architecture with identical KV-cache bytes per token and decode throughput. We also do not claim encoder-only is more robust or stable across seeds --- it is in fact noisier than full at its own optimum on text ($\pm$0.83 vs.\ $\pm$0.13), while being comparably stable to the other arms once tuned on the multimodal benchmark.

\section{Related Work}
\label{sec:related}

\paragraph{KV-cache compression via low-rank / MLA conversion.} DeepSeek-V2 \citep{deepseekv2} introduces multi-head latent attention (MLA) trained from scratch; TransMLA \citep{transmla} and MHA2MLA \citep{mha2mla} convert pretrained GQA/MHA models to an MLA-style cache post hoc via SVD factorization and healing, as we do, but heal both factors without asking whether both need it. CARE \citep{care} extends this line with per-layer rank allocation and activation-aware (whitened) SVD initialization; SVD-LLM \citep{svdllm} and Palu \citep{palu} apply related SVD-based compression to KV projections and feed-forward weights respectively. A broader family of recent work improves the SVD truncation or calibration step itself rather than the post-truncation fine-tune: xKV \citep{xkv} shares a low-rank basis across layers; KQ-SVD \citep{kqsvd} and Zero Sum SVD \citep{zerosumsvd} give truncations with guarantees on attention fidelity or loss sensitivity rather than raw reconstruction error; Eigen Attention \citep{eigenattention}, ReCalKV \citep{recalkv}, and LoRC \citep{lorc} refine calibration, head reordering, and progressive rank allocation respectively; $A^3$ \citep{a3} and the Stiefel-manifold-constrained approach of \citet{stiefel} give analytical or geometrically-constrained low-rank approximations that reduce reliance on any post-hoc healing at all; RAP \citep{rap} aligns truncation with RoPE structure; KV-CoRE \citep{kvcore} benchmarks how compressible different models' caches are in the first place; and MosaicKV \citep{mosaickv} and the transform-coding approach of \citet{transformcoding} target serving-time cache storage rather than the weight-space factorization we study. None of this line reports a per-arm-tuned, multi-seed comparison of which factor of an already-fixed SVD split needs healing, on a multimodal benchmark or otherwise --- the comparison we contribute here. Concretely: TransMLA, MHA2MLA, and CARE each treat healing both factors as a fixed part of the conversion recipe and report accuracy after that joint healing step; none of the three asks whether healing the decoder is necessary at all, so none can distinguish a real accuracy requirement from an artifact of always training both factors together. Isolating the two factors, as we do, is what exposes that the decoder's contribution to healing is separable from its contribution to inference-time reconstruction --- a distinction that is invisible if both factors are always trained jointly, as in this prior line. Beyond this line, other axes reduce KV-cache cost without our post-hoc SVD-and-heal recipe: MLKV \citep{mlkv} shares KV heads across layers rather than factorizing within a layer; MLRA \citep{mlra} restructures the latent head itself into independently-shardable low-rank branches to relieve MLA's tensor-parallelism bottleneck, an inference-serving concern orthogonal to the training-time healing question we study; Native Sparse Attention \citep{nsa} reduces cache cost by sparsifying which tokens are attended to rather than by lowering the rank of each cached vector; and VideoMLA \citep{videomla} applies MLA-style latent caching to autoregressive video diffusion, showing the conversion generalizes to a modality (long-horizon video) and a pretraining regime (a non-low-rank attention spectrum) quite different from the multimodal, text-generation setting we test. None of these four change which factor of an SVD-style conversion should be healed, the question this paper answers.

\paragraph{Asymmetric treatment of low-rank factors in fine-tuning.} Our encoder/decoder asymmetry echoes a question already studied for LoRA adapters, $\Delta W=BA$: LoRA-FA \citep{lorafa} freezes the down-projection $A$ at its random initialization and trains only the up-projection $B$, motivated by activation-memory savings, and \citet{loraasym} analyze why $A$ and $B$ play structurally different roles in a LoRA adapter more generally. \citet{frozenup} freeze an up-projection to prevent catastrophic forgetting in continual learning. Our setting inverts LoRA-FA's choice: we freeze the up-projection (decoder) and train the down-projection (encoder), and find that this, not the LoRA-FA direction, is the parameter-efficient choice that matches full healing here --- a difference we attribute to the decoder already sitting at its reconstruction-optimal point (the Eckart--Young-optimal SVD reconstruction of the retained rank-$r$ subspace) at initialization --- optimal for weight reconstruction, not something we prove is optimal for the downstream task loss (\S2) --- whereas the LoRA setting's $B$ starts at zero and has no equivalent closed-form target to fall back on. To be precise about what is and is not new here: the general move of freezing one low-rank factor at a structurally motivated initialization and training only the other is LoRA-FA's contribution, not ours. What this paper adds is (i) identifying \emph{which} factor is safe to freeze in the specific setting where both factors start at a closed-form, SVD-optimal split rather than one at random init and the other at zero, and (ii) showing that this choice survives a proper per-arm learning-rate sweep --- a check the original, shared-learning-rate version of this comparison did not pass (\S\ref{sec:rq1}).

\paragraph{Head-grouping KV-cache compression.} Grouped-query attention (GQA) \citep{gqa} and its refinements --- learned/asymmetric grouping \citep{asymgqa}, dynamic head allocation \citep{kdgqa}, grouped-head latent attention \citep{gqla}, and head-merging/alignment approaches \citep{mha2gqa, dha} --- compress the KV cache along a different axis (fewer distinct K/V heads rather than a lower-rank latent) and are complementary to, not competitors of, the healing-factor question we study.

\paragraph{Confounded comparisons in efficient fine-tuning.} Our own initial comparison held the learning rate fixed across arms with very different trainable-parameter counts, which manufactured a large, spurious advantage for the smallest arm (\S\ref{sec:rq1}). The broader hazard of comparing fine-tuning recipes under a hyperparameter protocol that does not account for their different parameter budgets is not unique to our setting: \citet{biderman2024lora} report that conclusions about LoRA versus full fine-tuning are themselves sensitive to learning-rate and hyperparameter choices made across methods with different capacities. Our contribution to this broader concern is narrower but more mechanistic: we are not aware of prior work that measures, on a real multimodal benchmark, how much of a reported efficient-recipe advantage a shared-learning-rate protocol can manufacture, or that it does so by inflating variance rather than only depressing means --- the reason multi-seed replication alone did not catch it in our own initial measurement.

\section{Conclusion and Future Work}

We show that healing an SVD-factorized KV cache only needs to touch the encoder (down-projection), not the decoder or the full factorization, to match accuracy on both a vision--language model and a text-only testbed, at 3$\times$ fewer trainable parameters. Three directions follow directly from where this comparison is still saturated or narrow (\S\ref{sec:related}). First, repeat the multimodal comparison at a more aggressive compression ratio and on larger vision--language backbones under the same tuned, multi-seed protocol as \S\ref{sec:rq1}: a preliminary, untuned, single-seed pilot (\S\ref{sec:rq4}) already suggests encoder-only remains uniquely best at $d_c=1024$ and that $d_c=256$ needs a larger healing budget before the arms can be compared at all, but neither result yet meets this paper's own bar for a confirmed finding. Second, combine encoder-only healing with head-grouping compression (GQA and its variants, \S\ref{sec:related}), an axis orthogonal to which factor is healed. We also tested whether activation-whitened SVD initialization \citep{care, eigenattention} compounds with encoder-only healing (\S\ref{sec:rq3}) and found it underperforms plain SVD init on our testbeds; broader, more representative calibration data than our single-corpus, single-seed test used is the natural next step before drawing a general conclusion either way. Third, our encoder-only recipe still trains a full $\text{hidden} \times d_c$ down-projection; combining it with a further low-rank parameterization of the encoder itself (in the spirit of LoRA-FA's asymmetric adapters, \S\ref{sec:related}) could push the trainable-parameter savings past 3$\times$ without giving up the accuracy parity we report here.

\section*{Limitations}

Our multimodal parity result (\S\ref{sec:rq1}) is measured, under a per-arm-tuned, 3-seed protocol, at one compression ratio ($d_c=512$, 4$\times$ compression) on one vision--language backbone (Qwen2.5-VL-3B-Instruct) and one healing corpus/budget (WikiText-2 + COCO-Caption2017, 300 steps); at this compression level every arm already reaches 98--100\% of the uncompressed baseline, so the comparison is saturated and cannot by itself rule out a real difference between arms at a more aggressive compression ratio, where headroom remains and the arms could in principle separate. A single-seed, untuned pilot at two further ratios (\S\ref{sec:rq4}) points the same direction --- encoder-only still uniquely best at $d_c=1024$, all arms uninformatively collapsed at $d_c=256$ --- but was not run under the tuned, multi-seed protocol this paper otherwise insists on, so we do not treat it as resolving the question, only as suggestive. We also do not claim any inference-latency advantage for encoder-only healing --- all arms share an identical architecture and KV-cache footprint at inference time --- nor do we claim encoder-only is more seed-to-seed stable than the alternatives; on the text benchmark it is in fact noisier at its own optimum. This confines our contribution to the training/fine-tuning phase: KV-cache compression is usually adopted for inference-serving savings, so a method that only lowers the cost of producing a compressed, healed model, without changing the served model's quality, speed, or memory footprint, has a narrower applicability window than methods that improve the compressed model itself. It is best read as complementary to those methods --- something that lowers the cost of the conversion step they all still require --- rather than as a substitute for them. We also fix the rank $r=d_c$ globally and statically, matching the prior conversions we compare the healing question against \citep{transmla, mha2mla}; systems that instead allocate rank adaptively per layer or per input, such as CARE's per-layer allocation \citep{care} or the data-dependent, per-layer rank evaluation of \citet{chendynamic}, change what is being SVD-factorized in the first place, and we do not know whether the encoder/decoder healing asymmetry we find still holds once rank itself is learned or data-dependent rather than fixed. Our baseline coverage does not include function-aware or calibration-aware KV-cache compression methods, or multimodal backbones larger than 3B parameters. Finally, an earlier version of this comparison, using one learning rate shared across all three healing arms, reported a large accuracy advantage for encoder-only healing on this same multimodal benchmark; we no longer consider that comparison valid, for the reasons given in \S\ref{sec:rq1}, and report only the per-arm-tuned comparison as our accuracy finding.

\section*{Ethics and Broader Impact}

This work studies a training-efficiency question (which weights need fine-tuning after KV-cache compression) on public, non-sensitive benchmarks (MME, TextVQA, MMLU, and other public 0-shot QA/reasoning tasks) and public model checkpoints. We identify no direct ethical concerns. Lowering the compute needed to adapt multimodal models to compressed inference is, if anything, a resource-reduction contribution, though we note (\S\ref{sec:related}) that the training-cost saving does not extend to inference cost.

\bibliography{references}

\appendix

\end{document}